\documentclass[letterpaper, 10 pt, conference]{ieeeconf}  
\usepackage{graphicx}
\usepackage{amsmath,amssymb,amsfonts}

\IEEEoverridecommandlockouts                              

\title{\LARGE \bf
Zero-Shot Sim-to-Real Reinforcement Learning for Fruit Harvesting
}

\author{Emlyn Williams$^{1}$ and Athanasios Polydoros$^{1}$
\thanks{$^{1}$Lincoln Centre of Autonomous Systems (L-CAS), School of Engineering and Physical Sciences, University of Lincoln, United Kingdom. Email: \{emwilliams, apolydoros\}@lincoln.ac.uk}%
\thanks{This work was supported by the Engineering and Physical Sciences Research Council [EP/S023917/1] and Dogtooth Robotics.}%
}

\begin{document}

\maketitle
\thispagestyle{empty}
\pagestyle{empty}

\begin{abstract}

This paper presents a comprehensive sim-to-real pipeline for autonomous strawberry picking from dense clusters using a Franka Panda robot. Our approach leverages a custom Mujoco simulation environment that integrates domain randomization techniques. In this environment, a deep reinforcement learning agent is trained using the dormant ratio minimization algorithm. The proposed pipeline bridges low-level control with high-level perception and decision making, demonstrating promising performance in both simulation and in a real laboratory environment, laying the groundwork for successful transfer to real-world autonomous fruit harvesting.

\end{abstract}

\setlength{\tabcolsep}{2pt} 

\section{INTRODUCTION}

Despite significant advances in robotics in recent years, manipulating objects in unstructured environments remains a challenging task. With marked variability in lighting, weather and plant geometries, agricultural environments present a notable challenge for robots to operate in. Agricultural tasks such as fruit harvesting require robots to operate under these variable conditions, all of which contribute to the difficulty of achieving reliable performance in the real world. Although some commercial robotic harvesting systems can successfully harvest up to 90\% \cite{pyman2024robots} of strawberries, they struggle with occluded fruit and dense clusters where the stem of the fruit is difficult to access. Further, unripe fruit surrounding the ripe target often results in missed detections or failed grasping. Whilst a human would naturally push an occluding fruit out of the way to collect the target strawberry (ensuring not to damage the unripe fruit), teaching a robot this behavior is challenging. Furthermore, fruits usually grow in close proximity within their clusters and thus they may be dynamically coupled, whereby the manipulation of one fruit within the cluster changes the state of adjacent fruits. This complicates efforts to grasp the stem without inflicting damage to the strawberry.

\begin{figure}[htbp]
    \centering
    \includegraphics[width=0.85\linewidth]{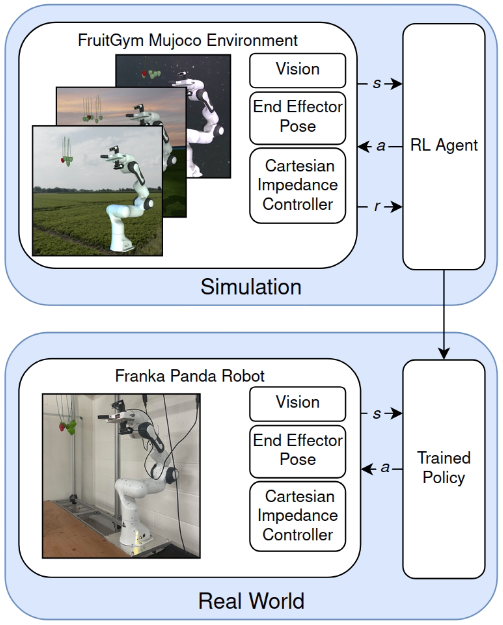}
    \caption{System overview illustrating our sim-to-real pipeline for robotic fruit harvesting. In the top panel (Simulation), the RL agent is trained in the custom FruitGym MuJoCo environment, receiving visual and pose observations $s$, reward $r$ and outputting control actions $a$. A Cartesian impedance controller then applies these actions at high frequency in simulation. Once trained, the same policy is deployed on the real Franka Panda robot (bottom panel), where the robot’s vision and end-effector pose are fed back to the trained policy for closed-loop control.}
    \label{fig:overview}
\end{figure}

Recent advances in sample-efficient reinforcement learning (RL) have enabled training of RL agents directly in the real world \cite{luo2024serl}\cite{luo2024precise}. However, the tasks demonstrated in these papers are factory-based tasks such as cable routing, RAM insertion and object relocation, with the environment being relatively consistent between episodes. Since we aim to enable fruit picking in the field, it is possible that training in one location with this approach would not generalize to other locations. To overcome this challenge, we leverage a sim-to-real pipeline.

Sim-to-real transfer techniques have emerged as a viable strategy to overcome these hurdles by enabling robust policy learning in simulation environments that can generalize to real-world conditions. In this work, we propose a novel sim-to-real system, illustrated in Fig. \ref{fig:overview}, for the autonomous harvesting of strawberries using a Franka Panda robot. Our approach leverages a custom-built Mujoco Gymnasium \cite{todorov2012mujoco}\cite{towers2024gymnasium} simulation environment specifically designed for strawberry picking, where real-world uncertainties are mimicked through extensive domain randomization. This simulation allows us to systematically vary lighting conditions, object placement and sensor noise, thereby preparing control and perception pipelines for the challenges encountered during actual robotic harvesting. Our contributions include:

\begin{itemize} 
    \item Fruit Gym - an open source Mujoco Gymnasium simulation environment tailored for a strawberry picking task, which incorporates realistic dynamics and environmental perturbations. Code available at \footnote{https://github.com/emlynw/fruit-gym}.
    \item Integration of domain randomization techniques to simulate real-world uncertainties such as variable illumination, occlusion and the spatial clustering of strawberries. 
    \item An RL training pipeline employing the dormant ratio minimization (DRM) algorithm that optimizes policy learning under stochastic conditions, with an emphasis on addressing failures in occluded and clustered fruit scenarios. 
    \item A comprehensive analysis of the proposed system's  performance and comparison with a baseline RL method.
\end{itemize}

The aforementioned contributions address robotic manipulation challenges for agricultural harvesting tasks, such as generalization to different operating environments and picking from dense clusters.

\section{RELATED WORK}

\subsection{Sim to Real transfer in Reinforcement Learning}

Training reinforcement learning (RL) agents in simulation and deploying them in the real world is a common strategy to avoid the expense and risk of real-world training. However, the \textit{sim-to-real} gap - discrepancies in sensor data and dynamics between simulation and reality - can cause learned policies to perform poorly on physical robots. A variety of methods have been proposed to overcome the sim-to-real gap \cite{zhao2020sim}. One such method is domain randomization, which randomizes various aspects of the simulation (e.g. textures, lighting and physics parameter) during training so that the policy learns to be robust to these factors, enabling zero-shot transfer to the real world \cite{chen2021understanding}. For example, Tobin et al. \cite{tobin2017domain} trained vision based networks in simulation with random textures and lighting to achieve successful real world robotic perception and control. Other sim-to-real techniques include domain adaptation, fine-tuning with a small amount of real data, meta-learning and using prior demonstrations \cite{zhao2020sim}. Domain randomization has, however, emerged as a simple and effective baseline for sim-to-real transfer in robotics. In the context of agricultural robotics, domain randomization is especially useful given the high variability in outdoor environments (e.g. lighting changes, varying crop appearances, weather conditions). By randomizing both visual inputs and physical parameters in our simulator (for instance, lighting, object positions and adding sensor noise), we aim to train a policy that generalizes across the diverse conditions the robot will encounter on real farms. The system proposed in this paper is the first, to the best of the authors' knowledge, that applies the concept of domain randomization for robot learning of harvesting tasks.

\subsection{Reinforcement Learning for Agricultural Robotics}

Whilst Learning from Demonstration (LfD) approaches such as behavioral cloning could be utilized for this application, these methods need a large number of demonstrations to converge \cite{zhao2020sim}. To avoid the time consuming process of collecting demonstrations in the field, we choose to utilize RL. The high variability of the environment, as previously mentioned, could also lead to a policy that does not generalize beyond the demonstration environment employed by LfD approaches \cite{zhao2020sim}. 

The application of RL to agricultural robotics is an emerging research area. Prior work in agriculture has begun to explore RL for tasks including harvesting and crop handling, though the literature is still relatively sparse compared to other domains. Yandún et al. (2021) applied deep RL to a grapevine pruning robot: using proximal policy optimization (PPO), they trained a 7-DoF manipulator to reach pruning points within vine canopies \cite{yandun2021reaching}. Notably, the learned policy achieved performance on par with a classical motion planning baseline (RRT in MoveIt) for selecting and reaching cut points. Their RL policy was not vision based, but relied on the input from a Faster-RCNN based perception pipeline which fed the end effector target position as part of the RL state space. Similarly, Lin et al. \cite{lin2022inverse} demonstrate a banana harvesting pipeline in which the target position is part of the RL state space. However, their work focuses on learning to solve the inverse kinematics (IK) of the robot through RL. Due to their reliance on the perception pipeline, such approaches may struggle with occlusions.

Beyond direct agricultural examples, inspiration can be drawn from RL successes in analogous manipulation tasks under field-like uncertainty. For instance, deep RL has been used for complex robotic manipulation in cluttered or deformable environments \cite{deng2019deep}\cite{matas2018sim}, indicating the potential to handle the variability inherent in farms and orchards. Whilst these works demonstrate growing interest in leveraging RL for agriculture, a fully end-to-end sim-to-real RL solution for robotic harvesting has yet to be explored. In most prior systems, perception and control are handled by separate modules, with computer vision algorithms detecting and localizing the target and a controller executing a predefined motion plan \cite{tang2020recognition}.  In contrast, our approach learns an integrated policy directly from raw sensory inputs (camera pixels and proprioception). To our knowledge, this is one of the first efforts to apply sim-to-real deep RL to an agricultural robot in an end-to-end manner, without a hand-engineered perception pipeline.

\section{Simulation Environment}\label{section:simulation}
Our sim-to-real pipeline for robotic strawberry picking comprises three principal components: a simulation environment, a low-level control module based on Cartesian impedance, and a high-level policy learning framework. In the following, we detail each component and describe how they interact.

\begin{figure}[htbp]
    \centering
    \begin{tabular}{ccc}
        \includegraphics[width=0.3\linewidth]{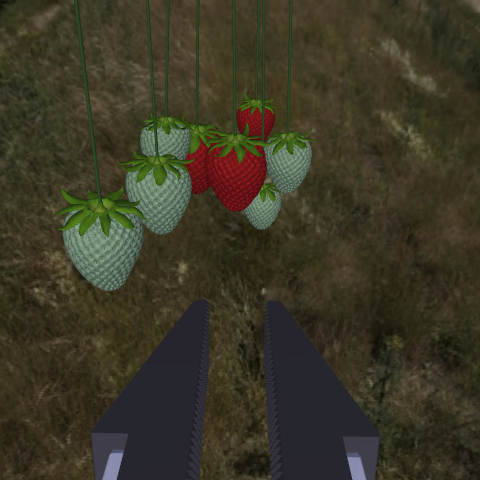} & 
        \includegraphics[width=0.3\linewidth]{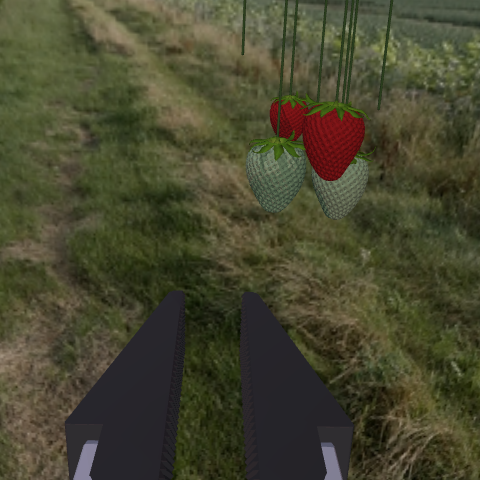} & 
        \includegraphics[width=0.3\linewidth]{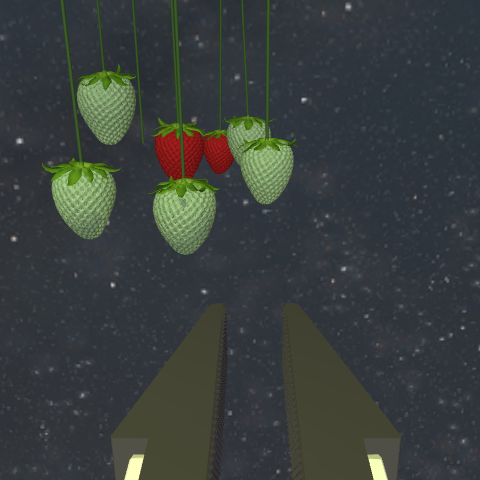} \\
        \includegraphics[width=0.3\linewidth]{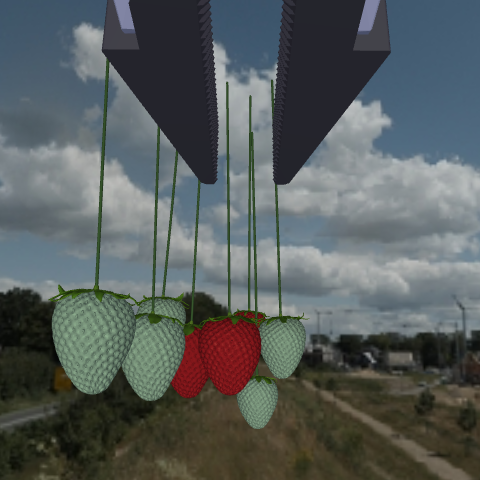} & 
        \includegraphics[width=0.3\linewidth]{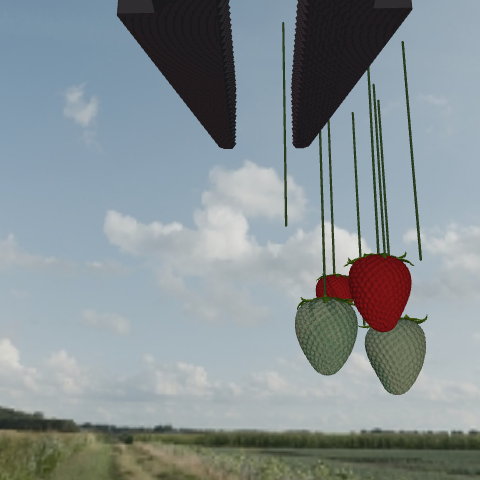} & 
        \includegraphics[width=0.3\linewidth]{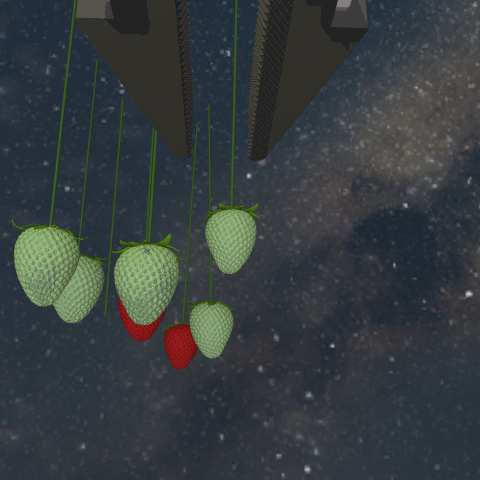} \\
        \includegraphics[width=0.3\linewidth]{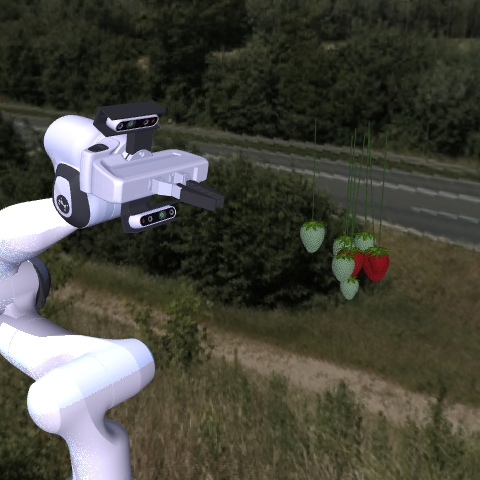} & 
        \includegraphics[width=0.3\linewidth]{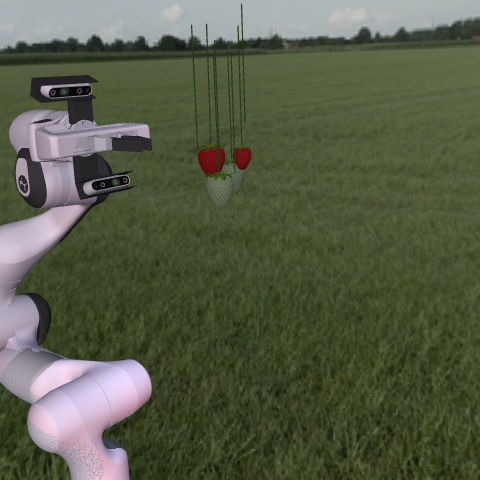} & 
        \includegraphics[width=0.3\linewidth]{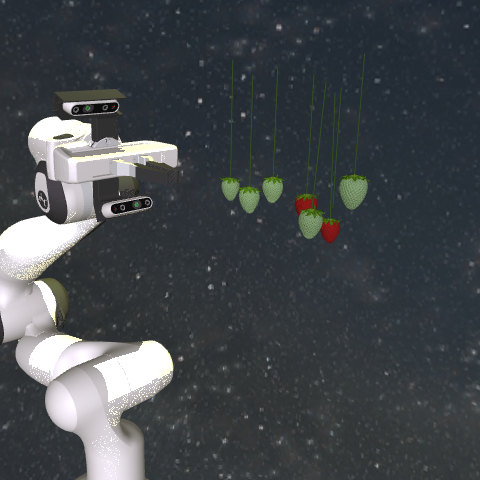} \\
    \end{tabular}
    \caption{FruitGym environments with domain randomization.}
    \label{fig:domain_randomization}
\end{figure}

We developed a custom simulation environment using the Mujoco physics engine and integrated it into the Gymnasium framework. The environment is designed to emulate a realistic strawberry picking scenario using a Franka Panda robot and supports extensive domain randomization. Two variants have been implemented:

\textbf{PickStrawbEnv} – In this environment, the robot is tasked with reaching for and grasping the stem of a single red strawberry target among multiple distractor (green) strawberries. Key features include:
\begin{itemize} 
    \item \textbf{Action Space:} A continuous 7-dimensional vector representing the change in three positional displacements, three rotational changes and a grasp command. The grasp command is interpreted as an attempt to grasp when above a threshold, and to release when below. The reason for using a discrete command for the action is due to the Franka Panda gripper sometimes taking up to 0.6 seconds to complete an action. When a gripper changes state from opened to closed (or vice versa), we advance the simulation until the action is complete to maintain the Markov properties of the environment \cite{luo2024precise}. 
    \item \textbf{Observation Space:} A dictionary containing a state vector and image observations. The state vector includes the end-effector pose (position and quaternion orientation), velocity, gripper position, and a gripper state vector. The image observations include feeds from two wrist cameras which provide RGB images. 
    \item \textbf{Domain Randomization:} Variability is introduced via random perturbations in lighting (position, intensity and headlight properties), camera parameters (position and orientation noise) and object positions, thus emulating real-world uncertainties. Examples of domain randomization are shown in figure \ref{fig:domain_randomization}. 
\end{itemize}

\textbf{PickMultiStrawbEnv} – This variant extends the single-target setup to multi-target scenarios. When a red strawberry is successfully grasped, it is removed from the environment; prompting the robot to focus on remaining targets. This environment further challenges RL algorithms with multi-object manipulation tasks.

\subsection{Low-Level Control via Impedance Controller}\label{controller}

Our low-level control module is adapted directly from the SERL framework \cite{luo2024serl}. In this architecture, a two-layer control hierarchy is employed: a high-level RL policy issues actions at a low frequency (we run the policy at 20Hz), with a real-time impedance controller tracking these actions at 1kHz. The controller’s objective is formulated as
\[
F = k_p \cdot e + k_d \cdot \dot{e} + F_{\text{ff}} + F_{\text{cor}},
\]
where \(e = p - p_{\text{ref}}\) is the error between the measured pose \(p\) and the reference pose \(p_{\text{ref}}\), \(F_{\text{ff}}\) is the feed-forward force, and \(F_{\text{cor}}\) accounts for Coriolis effects. These computed forces are converted into joint torques via the Jacobian transpose, with additional nullspace torques ensuring joint stability. A key aspect of the SERL approach is the clipping of the reference error so that \(|e| \leq \Delta\). This constraint bounds the interaction force, ensuring safe operation during contact-rich tasks.

\section{Policy Learning}

We adopt a reinforcement learning pipeline built upon the Dormant Ratio Minimization (DRM) algorithm \cite{xu2023drm}. DRM builds upon DrQv2 \cite{yarats2021mastering}, with modifications aimed at improving sample efficiency and reducing sustained network inactivity early in learning. DRM mitigates the common issue of network inactivity - quantified by the dormant ratio, which measures the fraction of nearly inactive neurons—by explicitly minimizing this metric through three complementary mechanisms:

\begin{itemize}
    \item \textbf{Dormant-Ratio-Guided Perturbation:} Periodically perturbing the network weights based on the current dormant ratio to rejuvenate network expressivity and encourage exploration.
    \item \textbf{Awaken Exploration Scheduler:} Adapting the exploration noise dynamically so that when the dormant ratio is high, the agent receives large noise to encourage exploration. As the ratio decreases, the noise is reduced to favor exploitation.
    \item \textbf{Dormant-Ratio-Guided Exploitation:} Adjusting the value target by incorporating a dormant-ratio-dependent parameter, enabling the agent to better exploit past successful experiences when the dormant ratio is low.
\end{itemize}

In our implementation, we expand the original DRM framework to accept inputs from two cameras, with one camera placed below the gripper to view the strawberry and another above the gripper to view the stem. A separate encoder is trained for each image with the outputs of both encoders being concatenated before being passed to the actor and critic modules.

\subsection{Reward Function}

For the PickStrawbEnv which contains one target (red) strawberry, assume:

\begin{itemize}
    \item $\mathbf{x}_{r}$ is the position of the red strawberry. For the PickStrawbMultiEnv, $\mathbf{x}_{r}$ is the position of the nearest red strawberry to the end effector. 
    \item $\mathbf{x}_{tcp}$ is the end effector tool center point (TCP) position.
    \item $\mathbf{x}^{t}_{g}$ is a vector of the green strawberry positions at timestep $t$.
    \item $\mathbf{a}$ is the action vector and $\mathbf{a}_{\text{prev}}$ is the previous action.
\end{itemize}

The reward function is a combination of the following factors:

\textbf{Grasp Reward:}
A binary reward given for successfully grasping the stem of the target strawberry and only the target strawberry. This is implemented by monitoring collisions between the inner face of both of the gripper fingers and other objects.
\begin{equation}
r_{\text{grasp}} =
\begin{cases}
1, & \text{if both fingers contact the target stem} \\
0, & \text{otherwise}
\end{cases}
\end{equation}

\textbf{End Effector-Target Proximity Reward:}
This rewards the agent for moving the gripper towards the desired picking point on the stem of the strawberry.
\begin{equation}
r_{\text{prox}} = 1 - \tanh\Bigl(5\,\|\mathbf{x}_{r} - \mathbf{x}_{tcp}\|\Bigr)
\end{equation}

\textbf{Displacement Penalty}
The penalizes the agent for moving the green strawberries away from their initial position. The aim of this penalty is to minimize damage to unripe strawberries.

\begin{equation}
    r_{\text{green}} = 1 - \tanh\Biggl(5\,\sum_{i} \Bigl\|\mathbf{x}_{g,i}^{\text{t}} - \mathbf{x}_{g,i}^{t_{0}}\Bigr\|\Biggr)
\end{equation}

Where $t_0$ is the initial timestep of the episode, and $N$ is the number of green strawberries.

\textbf{Energy Penalty:}
This penalizes the agent for taking large actions, encouraging it to move precisely.
\begin{equation}
r_{\text{e}} = -\|\mathbf{a}\|
\end{equation}

\textbf{Smoothness Penalty:}
This penalizes the agent for changing directions quickly, in order to minimize accelerations. 
\begin{equation}
r_{\text{s}} = -\|\mathbf{a} - \mathbf{a}_{\text{prev}}\|
\end{equation}

The overall reward is given by:
\begin{equation}
R= w_{\text{grasp}}\,r_{\text{grasp}} + w_{\text{prox}}\,r_{\text{prox}} + w_{\text{green}}\,r_{\text{green}} + w_{e}\,r_{\text{e}} + w_{s}\,r_{\text{s}}
\end{equation}

where $w_{\text{grasp}}=8$, $w_{\text{prox}}=4$, $w_{\text{green}}=1$, $w_{e}=2$, $w_{s}=1$. 

\section{Real Robot setup}

We use the Franka Panda robot using a ROS 2 implementation of the Cartesian impedance controller described in section \ref{controller}. The image inputs are given by two Intel Realsense d435 cameras on 3d printed mounts on the robot end effector.

For our strawberry picking tests, we use plastic strawberries with green wire as stems/vines, with metal washers attached at the end of the stems. For quick randomizations between resets, we hang the strawberries using magnets so that the location and order of the strawberry can be quickly changed during resets. We will test on real strawberry plants when the picking season begins.

\section{Experiments}

We compare the DRM algorithm to a DrQv2 baseline, training for 2 million timesteps for 3 random seeds per algorithm. The state space consists of two wrist cameras, end effector pose, and gripper width. For all experiments we use the PickStrawbEnv (one target strawberry) with all of the domain randomization methods described in section \ref{section:simulation}. The number of strawberries was randomized each episode, with a maximum of 7 greens in the scene, and a minimum of 1.

To evaluate the performance of the learned policy, we measure the success rate of the policy while varying the number of green strawberries in the scene both in simulation and in the real world. We run $180$ trials for both the simulation and real environments, consisting of $30$ trials for six configurations, with the positions of the strawberries being randomized for each trial. In simulation, an episode is counted as a success if the agent receives the grasp reward (described in section \ref{section:simulation}) during the episode. For the real world experiments, the robot operator determines the episode a success if the gripper successfully grasps the target strawberry stem, and nothing else. An episode is deemed a failure if the agent grasps a green strawberry stem or does not grasp the target strawberry within 300 steps (15 seconds). 

\section{Results and Discussion}

\begin{figure*}[htbp]
    \centering
    \includegraphics[width=0.16\linewidth]{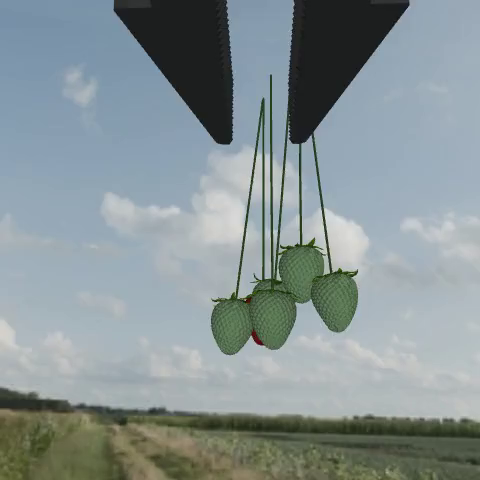}%
    \includegraphics[width=0.16\linewidth]{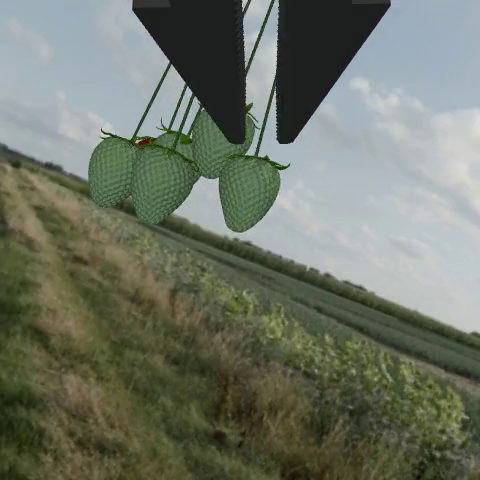}%
    \includegraphics[width=0.16\linewidth]{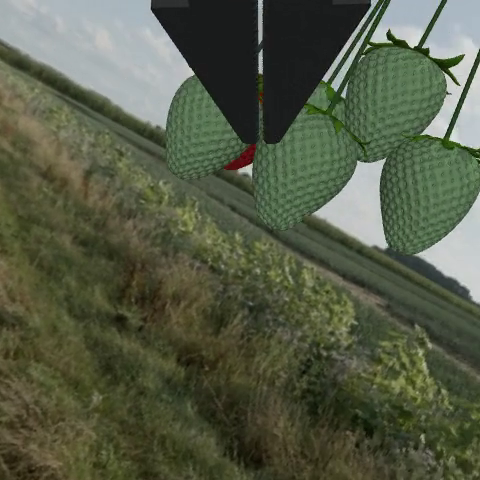}%
    \includegraphics[width=0.16\linewidth]{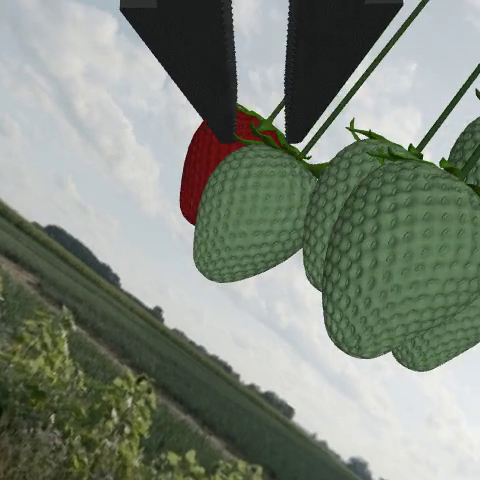}%
    \includegraphics[width=0.16\linewidth]{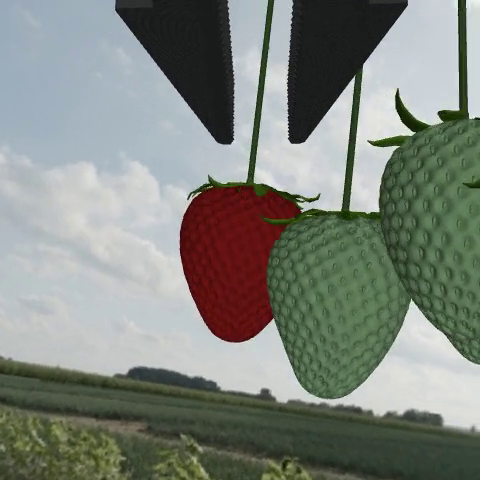}%
    \includegraphics[width=0.16\linewidth]{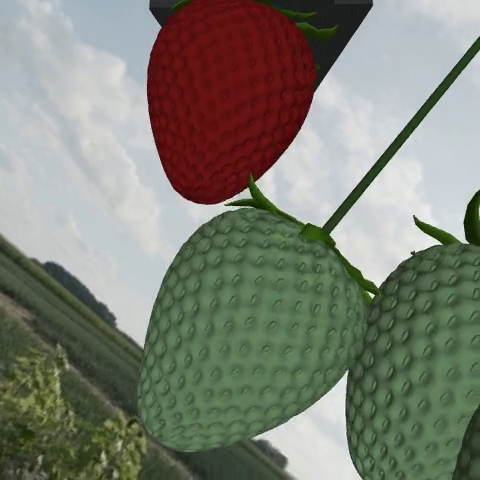}%
    \\[0.0em] 
    \includegraphics[width=0.16\linewidth]{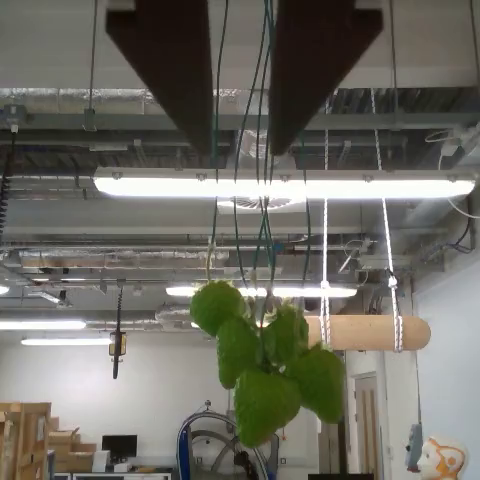}%
    \includegraphics[width=0.16\linewidth]{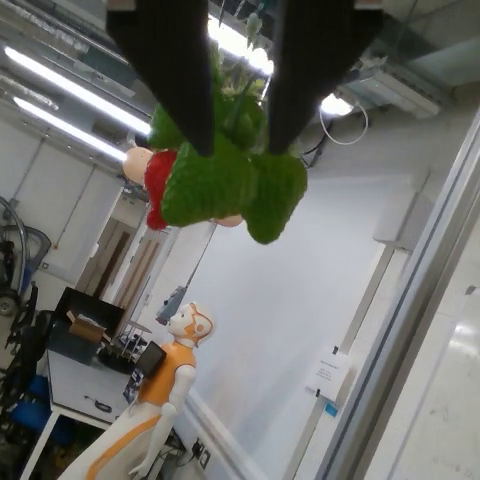}%
    \includegraphics[width=0.16\linewidth]{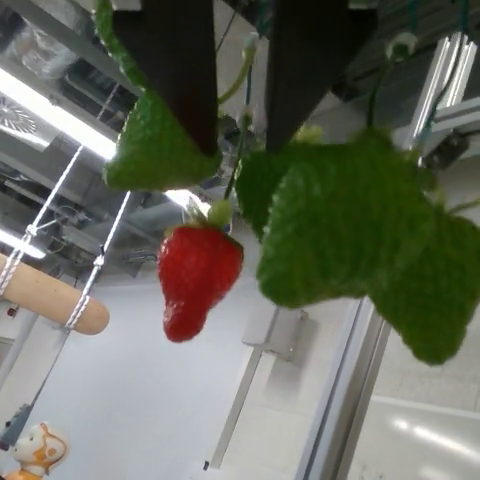}%
    \includegraphics[width=0.16\linewidth]{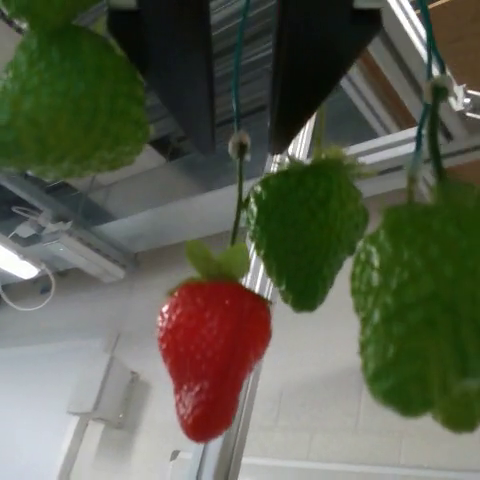}%
    \includegraphics[width=0.16\linewidth]{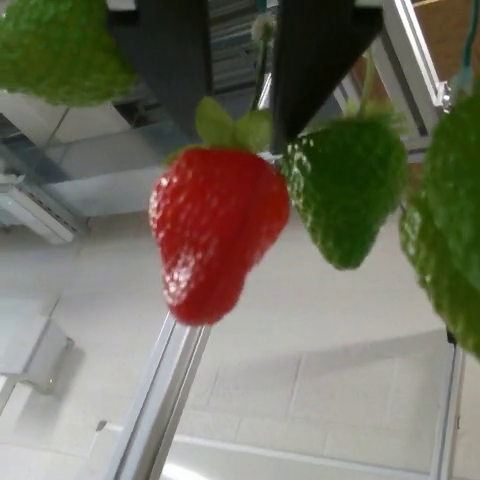}%
    \includegraphics[width=0.16\linewidth]{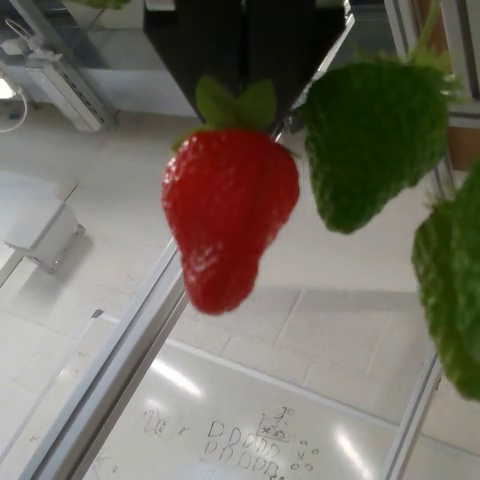}%
    \caption{Examples of successful attempts in sim and real environments in a cluster containing 5 green strawberries and 1 red.}
    \label{fig:demo}
\end{figure*}

\begin{figure*}[htbp]
    \centering
    \includegraphics[width=0.16\linewidth]{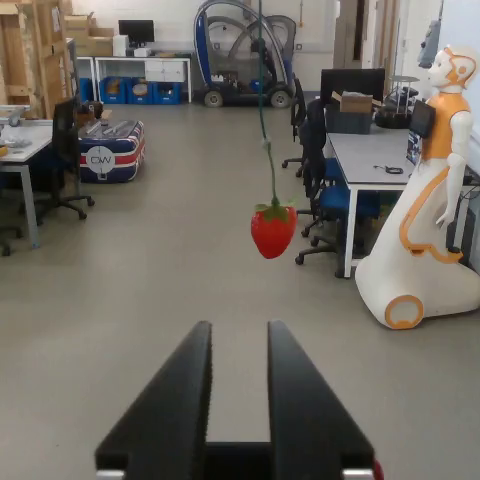}%
    \includegraphics[width=0.16\linewidth]{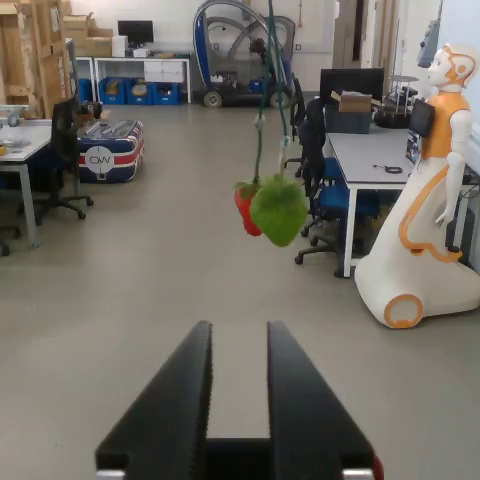}%
    \includegraphics[width=0.16\linewidth]{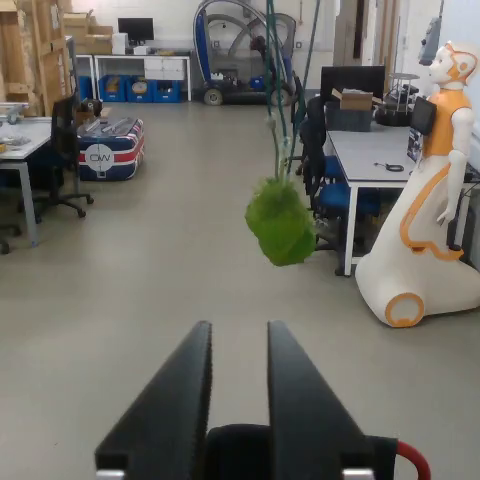}%
    \includegraphics[width=0.16\linewidth]{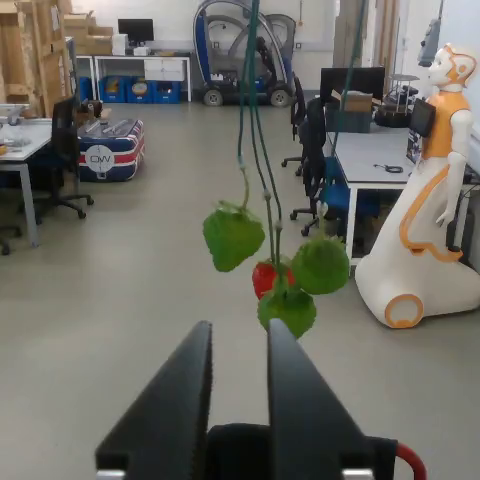}%
    \includegraphics[width=0.16\linewidth]{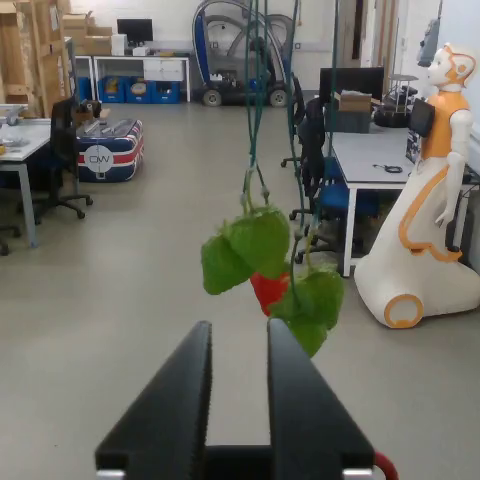}%
    \includegraphics[width=0.16\linewidth]{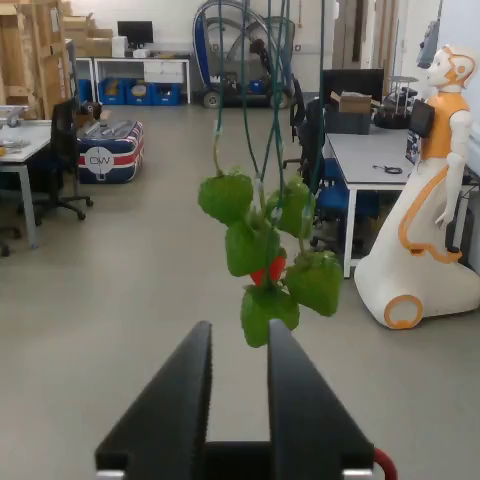}%

    \caption{Evaluation cluster examples with varying number of green strawberries.}
    \label{fig:clusters}
\end{figure*}

Figure \ref{fig:demo} shows examples of successes in both simulation and in the real world when harvesting from a complex cluster containing five green strawberries and one red. In the initial state, the red strawberry is only partially visible to one of the two wrist cameras. The agent learns to move towards the cluster and push between the stems of the green strawberries to grasp the stem of the red. As can be seen in the bottom half of figure \ref{fig:demo}, the real world experiment takes place in a lab environment with a cluttered background that has not been seen in training. Both examples are taken from the same DRM policy.

\begin{figure}[htbp]
    \centering
    \includegraphics[width=0.85\linewidth]{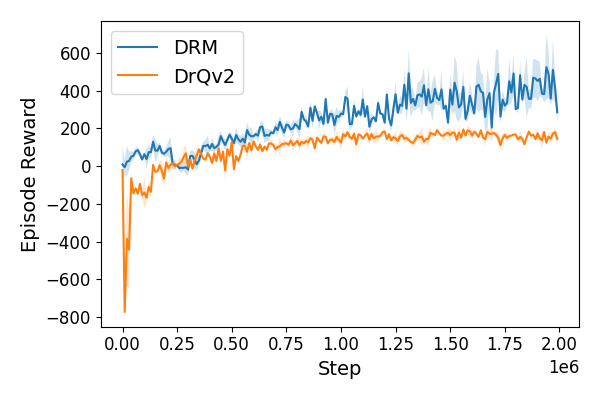}
    \caption{RL episode reward throughout training. Results are averaged over 3 seeds with a standard deviation shading of $\pm0.5$.}
    \label{fig:reward_comaprison}
\end{figure}

\begin{figure}[htbp]
    \centering
    \includegraphics[width=0.85\linewidth]{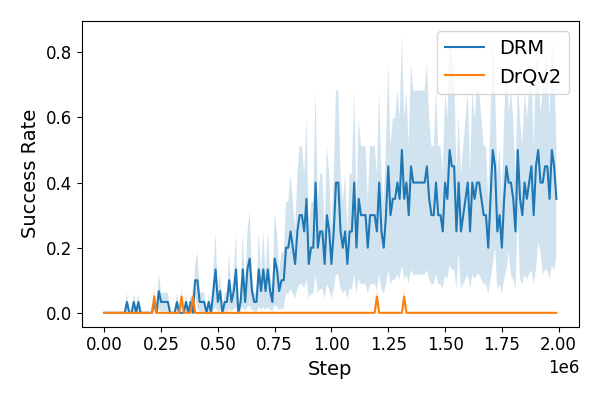}
    \caption{RL success rate throughout training. Results are averaged over 3 seeds with a standard deviation shading of ±0.5.}
    \label{fig:success_comparison}
\end{figure}

Figures \ref{fig:reward_comaprison} and \ref{fig:success_comparison} show graphs of episode rewards and success rates of the agent throughout training, respectively. Of the three DRM seeds trained, one learned to successfully grasp the strawberry stem, whilst the others would move the gripper close to the desired grasping point but not successfully grasp. The three DrQv2 seeds also moved towards the desired grasping point but did not learn successful grasping behavior. We believe that this difficulty in learning is due to the gripper operating with a discrete action, since both DRM and DrQv2 are designed as continuous action space algorithms.  

\begin{figure}[h!]
    \centering
    \includegraphics[width=0.85\linewidth]{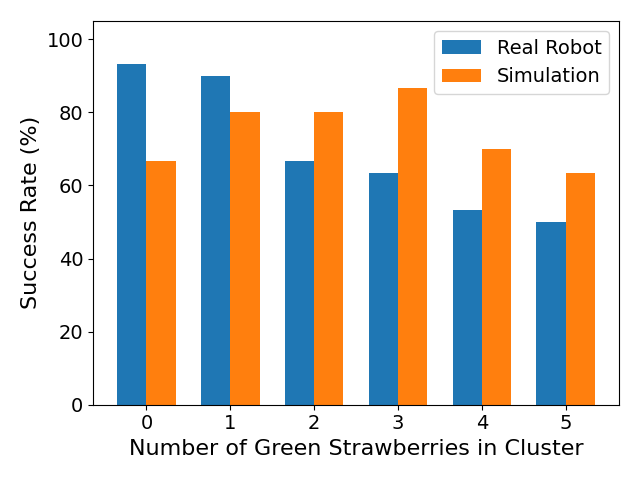}
    \caption{Success rate comparison of simulation and real robot. 30 trials were conducted for each number, with the positions of the strawberries being randomized for each trial.}
    \label{fig:bar_chart}
\end{figure}

Examples of the different evaluation configurations are shown in figure \ref{fig:clusters}. Figure \ref{fig:bar_chart} shows the success rate of the agent in both simulation and in the real world when grasping a single target strawberry with different numbers of green strawberries. When deployed on the real robot the success rate decreased as the number of green strawberries increased; with a success rate of 95\% with no green strawberries in the scene and 50\% with 5. This trend was not observed in simulation however, as the agent achieved the highest success rate with 3 green strawberries in the scene. With 0 and 1 green strawberries, the agent performed better in the real world than in simulation. Common failure modes included the end effector just missing the target and knocking it aside, giving it a velocity. This resulted in the agent trying to pick a moving target, thereby increasing the difficulty of the task. Another failure mode was the robot briefly grasping a green strawberry before grasping the red.

\subsection{Limitations}

\begin{itemize}
    \item \textbf{Temporal dependencies:} The network architecture used in this paper encodes temporal information by stacking subsequent states (we use a frame stack of 2). Increasing the frame stack results in a higher dimensionality of the state space, increasing the complexity of the problem. Integration of recurrent layers (such as LSTM) could improve performance due to improved handling of partial observability \cite{hausknecht2015deep}, such as in dynamic scenarios when the target is moving. This could also aid in exploration if the target is not captured in the observation. 
    \item \textbf{Action Space Mismatch:} As previously mentioned, the combination of the continuous robot position action combined with the discrete gripper action could harm performance. We believe that this is the reason for some seeds failing to learn successful grasping.\cite{luo2024precise} approach this issue by using a separate critic for the grasp action. 
    \item \textbf{Computational Demands:} The high fidelity of the simulation combined with the large state space given by using two camera inputs leads to significant computational demands, with 2 million learning steps taking 48 hours on an Nvidia A100 GPU and a replay buffer size exceeding 40GB for 1 million transitions.
\end{itemize}



\section{Future Work}

\textbf{Fine tuning in the real world:} We demonstrate very good performance in the real world, achieving a 50\% success rate on the challenging case of picking from a dense cluster of 5 green strawberries and 1 red. However, figure \ref{fig:bar_chart} shows that with more green strawberries, a gap remains between simulation performance and real world performance. Future work will focus on ways of fine tuning the policy in the real world using learned reward functions. 

\textbf{Expanding and improving FruitGym:} We will add other plants such as tomatoes to FruitGym and improve the performance of the simulation by adapting it to work with Mujoco Playground \cite{zakka2025mujocoplayground} which enables GPU parallelization.

\textbf{Real World Validation}
We will test our approach outside of the lab to pick real strawberries when the strawberry picking season begins. For this we will choose an appropriate tool for cutting the strawberry stem while grasping.

\section{CONCLUSION}

We have presented a novel sim-to-real pipeline that integrates a custom Mujoco simulation environment, Cartesian impedance control and a deep RL training framework with the DRM algorithm for agricultural harvesting tasks. The proposed method was evaluated in diverse conditions in simulation and a lab environment in the real world, achieving state of the art performance in strawberry picking from dense clusters. Our work lays the foundation for future research in autonomous agricultural robotics using sim-to-real transfer.

\addtolength{\textheight}{-12cm}   




\section*{ACKNOWLEDGMENT}

We use OpenAI's ChatGPT to edit parts of the paper for grammatical purposes.


\bibliographystyle{ieeetr}
\bibliography{references.bib}

\end{document}